\documentclass[smallextended]{svjour3}       
\renewcommand{\makeheadbox}{}

\smartqed  
\usepackage{graphicx}
\usepackage[hyphens]{url}
\usepackage[inline]{enumitem}
\usepackage{caption}
\usepackage{framed,multirow}
\usepackage{times}
\usepackage{textcomp}
\usepackage[numbers,sort&compress]{natbib}

\usepackage{fancyhdr}
\journalname{}
\begin{document}

\title{Where Is My Puppy? Retrieving Lost Dogs by Facial Features\thanks{This work was supported in part by FAPESP, CAPES, CNPq, and SAMSUNG.}
}


\author{Thierry Pinheiro Moreira \and
        Mauricio Lisboa Perez \and
        Rafael de Oliveira Werneck \and
        Eduardo Valle \thanks{T. Moreira, M. Perez, and R. Werneck contributed equally to this work.}
}


\institute{T. Moreira \at
              LIV Lab., Institute of Computing (IC), University of Campinas (Unicamp) \\
              \email{thierrypin@liv.ic.unicamp.br}           
           \and
           M. Perez \and R. Werneck \and E. Valle \at
              RECOD Lab., Institute of Computing (IC), University of Campinas (Unicamp)
           \and
           E. Valle \at
	      Department of Computer Engineering and Industrial Automation (DCA), School of Electrical and Computer Engineering (FEEC), University of Campinas (Unicamp) – Av. Albert Einstein, 400, Campinas 13083-852, SP, Brazil
}

\date{Received: date / Accepted: date}

\maketitle

\begin{abstract}

%
A pet that goes missing is among many people's worst fears: a moment of distraction is enough for a dog or a cat wandering off from home. 
Some measures help matching lost animals to their owners; but automated visual recognition is one that --- although convenient, highly available, and low-cost --- is surprisingly overlooked.
%
In this paper, we inaugurate that promising avenue by pursuing face recognition for dogs. 
We contrast four ready-to-use human facial recognizers (EigenFaces, FisherFaces, LBPH, and a Sparse method) to two original solutions based upon convolutional neural networks: BARK (inspired in architecture-optimized networks employed for human facial recognition) and WOOF (based upon off-the-shelf OverFeat features). 
Human facial recognizers perform poorly for dogs (up to 60.5\% accuracy), showing that dog facial recognition is not a trivial extension of human facial recognition. The convolutional network solutions work much better, with BARK attaining up to 81.1\% accuracy, and WOOF, 89.4\%. The tests were conducted in two datasets: Flickr-dog, with 42 dogs of two breeds (pugs and huskies); and Snoopybook, with 18 mongrel dogs.

\keywords{Face Recognition \and Dog Recognition \and Deep Learning \and Convolutional Networks}

\end{abstract}



\section{Introduction}


Dogs hold a special status for humans: they were the first domesticated animals, and are beloved as pet companions. The American Pet Products Association estimates that they are present in $65\%$ of U.S. households \footnote{\url{http://www.humanesociety.org/issues/pet_overpopulation/facts/pet_ownership_statistics.html} (December 2015).}.

However, our love for dogs becomes worry when they run away. The American Society for the Prevention of Cruelty to Animals\footnote{\url{https://www.aspca.org/about-us/press-releases/how-many-pets-are-lost-how-many-find-their-way-home-aspca-survey-has-answers} (March 2015).} reports that 15\% of pet-owning households lost a dog or a cat in the last five years; and 15\% of those lost animals were never recovered. Most dog owners recovered their dogs through searches in the neighborhood (49\%), or through ID tags (15\%). Only 6\% of owners found their pets at shelters.

Owners have a handful of ways to locate lost dogs: collar tags, GPS devices, tattoos, implanted microchips/RFIDs. Collar tags are cheap, but easily lost. In addition, not all dogs tolerate wearing a collar all the time. Tattoos and implanted chips are more robust, but also more expensive and require a preventive attitude by the owners. Some first-world countries solve that problem by making tattooing or chip implant mandatory. But in other countries, including Brazil, tattoos and implants are virtually unheard-of.

Thus the main motivation of this work: why not identifying lost dogs using... the dogs themselves? Contrarily to collars and IDs, the dog appearance is intrinsic. Contrarily to chips and implants, appearance is readily available and does not require extraordinary forethought. Most, if not all, owners will spontaneously accumulate dozens of pictures of their pets.

In this paper, we inaugurate that promising avenue by pursuing face recognition for dogs. Although literature on pet biometry is --- to say the least --- scarce, human face recognition is a well-established domain with a vast literature~\cite{Ahonen2004,Ahonen2006,Belhumeur1997,Chiachia2014,Cox2011,Pinto2011,Sirovich1987,Turk1991,Wiskott1997}. That allows us to take inspiration from methods intended for humans, and to evaluate their effectiveness for dogs.

\section{Literature review}
\label{review}


Literature on animal biometry is scarce. Computer Vision is mainly interested in animals as members of large classes, not as individuals. The interest in animals as individuals come mainly from wildlife researchers, for whom the identification of specimens is critical to determine ecological patterns\footnote{For example, in August 2015, the New England Aquarium and MathWorks have launched a large competition for identifying individual endangered whales: \url{https://www.kaggle.com/c/noaa-right-whale-recognition}.}.

The art on biometry for pets is virtually nonexistent. However, some works address fine-grained classification of dogs into breeds. Parkhi et al.~\cite{parkhi2012cats} used a model to describe both the shape and the texture of a pet in their Oxford-IIIT Pet dataset, achieving $59\%$ of accuracy. Oxford-IIIT Pet is a public and carefully annotated dataset\footnote{\url{http://www.robots.ox.ac.uk/~vgg/data/pets/}} containing $12$ cat breeds, and $25$ dog breeds, with $200$ images for each breed, downloaded from Google Images, Flickr, Catster, and Dogster. Unfortunately, each individual appears only once in this dataset, rendering it useless for our purposes. Parkhi et al.~\cite{parkhi2012cats} uses a deformable part model~\cite{felzenszwalb2010object} with HOG filters~\cite{dalal2005histograms} to detect edge information and represent shapes. They aim at detecting stable and distinctive components of the dog, particularly the head. Bag of visual words (from densely sampled SIFT descriptors~\cite{lowe1999object}) were used for texture.

Liu et al.~\cite{liu2012dog} classified dogs into breeds using models for the geometry and appearance of breeds, including facial clues. First, they detect the dog face using a sliding window, then locate the nose and eyes using consensus of exemplars~\cite{belhumeur2013localizing}, and finally infer four possible locations for the ears and remaining components. Next, they extract color histograms, and grayscale SIFT descriptors, from different parts of the dog face. They achieved $67\%$ accuracy in a database created by the authors with $133$ dog breeds and 8,351 images. Wang et al.~\cite{wang2014dog}, working in this same dataset, improved the results to $96\%$ accuracy, representing dog shapes using 8 landmarks given by points on the Grassmann manifold. Grassmann manifolds are differential geometry structures used, among other things, for image registration.


The only vision studies we could find that consider dogs as individuals are not about Computer Vision, but about the Human Visual System. Scapinello et al.~\cite{scapinello1970role} studied how familiarity and orientation affects recognition of human faces, canine faces, and buildings. Diamond and Carey~\cite{diamond1986faces} also study how 180-degree rotations affect the recognition of pictures, including dogs, by humans. Both works find that rotations affect the recognition of faces, but Scapinello et al.~\cite{scapinello1970role} show that the effect is greater for human faces than for dog faces. More important to us is that Diamond and Carey~\cite{diamond1986faces} report only 76\% accuracy for humans identifying dogs in optimal conditions. When the subjects were experts (breeders and judges from the American Kennel Club/New York) the accuracy raised to 81\%. Those numbers suggest that recognizing a large number of dogs is far from trivial, even for human experts.


Human Face Recognition is a well-established topic~\cite{Ahonen2006,Belhumeur1997,Turk1991,Wiskott1997}, with different paradigms: \begin{enumerate*}[label=\itshape\roman*\upshape)] \item \emph{detection}, in which the system detects whether a face is present in the image (i.e., any face, not someone's face in particular); \item \emph{verification}, in which the system decides whether two faces come from the same person; and \item \emph{identification}, where the system matches an unknown face against a labeled set~\cite{Pinto2011}. \end{enumerate*}

Detection is useful as pre-processing for the other paradigms, and has application of its own, e.g., in camera auto-focusing. Verification is useful, for example, for biometric authentication. However, here we will focus on identification.

Sirovich and Kirby~\cite{Sirovich1987} first introduced \emph{Eigenfaces} for face detection. It employs the eigenvectors of the image; hence the name. In Eigenfaces, one starts with a large collection of face images, all of the same size, and all (approximately) registered. The images are taken as vectors, whose covariance matrix is estimated. The eigenvectors of that covariance matrix are computed, and those corresponding to the highest eigenvalues are kept to form a subspace in which the faces are represented (the whole procedure is akin to a Principal Component Analysis). Soon, Eigenfaces became a tool for verification and identification as well~\cite{Turk1991}. 

Fisherfaces are similar to Eigenfaces, but using Fisher Linear Discriminants~\cite{Fisher1936}, or, very often Linear Discriminant Analysis (although each technique gives slightly different results). A subspace is chosen in order to minimize within-class distances and maximize between-class distances~\cite{Belhumeur1997}. While Eigenfaces emphasizes the ability to reconstruct the original image using a small number of components, Fisherfaces emphasizes the ability to discriminate faces from different people.

Contrasting with those holistic approaches, Local Binary Patterns describe small patches of the image. Surrounding pixels are thresholded against the center of the neighborhood, resulting in a compact bit array. The local binary patterns can be pooled together into a single histogram per image~\cite{Ahonen2006}. At first, it was used for texture classification; later, it was applied to different domains, such as face recognition.

Newer approaches use sparse coding for face recognition. Xu et al.~\cite{xu2011} presented a two-phase method. First, the query image is represented as a linear combination of the $n$ training faces: $y = a_1x_1 + \dots + a_nx_n$, being y the reconstructed image, and $a_i$ and $x_i$ the \textit{i}th coefficient and train image. The contribution of the \textit{i}th face to the reconstruction is taken as $a_ix_i$, and its deviation to the reconstruction, $e_i = ||y - a_ix_i||^2$, is the measure of distance to select the $M$ nearest faces.

The second phase is, again, reconstructing the query image, now using only the selected $M$ faces. Between the selected images, there is a subset of the total classes. The contributions of all elements of the set are summed to obtain the class contribution to the reconstruction. Finally, the class with the smallest deviation to the reconstruction, computed as before, is chosen as the verdict of the system.


Convolutional Neural Networks have seized the attention of the Computer Vision community in virtually all tasks, and facial recognition is no exception.

Pinto et al.~\cite{Pinto2011} evaluated networks with two to three layers, which extract features that are then fed to Support Vector Machines, in a one-versus-all configuration. They optimized their networks by trying a range of network architectures initialized with random weights. In this work the weights themselves are not optimized at all. They achieved 85\texttildelow89\% identification accuracy with the best network architecture on the public datasets Facebook100 and PubFig83.

Chiachia et al.~\cite{Chiachia2014} improved that result by tailoring the last layer of the network for each individual, and optimizing both the hyperparameters and the weights of that layer. They raised the state of the art on PubFig83 to 92\% accuracy.

Those networks, although successful, are still rather shallow, with up to three layers.
Deeper Convolutional Neural Networks have obtained state-of-the-art results in many image classification tasks~\cite{Krizhevsky2012,Sermanet2013,Szegedy2014}. Although those network are very expensive to train --- both in computational resources and in training samples --- it is often possible to reuse the weights learned from a task to another, in what is called transfer learning.

The ImageNet Competition is a challenging annual contest to find the best classification model for a thousand different classes, using a training set of over 1.2 million images. Since 2012, Deep Convolutional Networks  got systematically the best places. In particular, Sermanet et al.~\cite{Sermanet2013}, improving on the winner model for ImageNet 2012~\cite{Krizhevsky2012}, proposed a model who won the localization task for ImageNet 2013. They distributed a ready-to-use model, with the weights already learned, as OverFeat\footnote{\url{http://cilvr.nyu.edu/doku.php?id=software:overfeat:start} (August 2015)}. OverFeat became, through transfer learning, a very popular model for feature extraction in a myriad of tasks beyond ImageNet.

\section{Material and methods}
\label{sec:material_methods}


\subsection{Datasets}
\label{sec:datasets}

We employed two datasets, described below.

\subsubsection{Flickr-dog}
We acquired the Flickr-dog dataset\footnote{available at: 
To be published on article acceptance.} by selecting dog photos from Flickr 
available under Creative Commons licenses. We cropped the dog faces, rotated them to align the eyes horizontally, and resized them to $250\times250$ pixels.

We selected dogs from two breeds: pugs and huskies. Those breeds were selected to represent the different degrees of challenge: we expected pugs to be difficult to identify, and huskies to be easy. For each breed, we found 21 individuals, each with at least 5 photos. We labeled the individuals by interpreting picture metadata (user, title, description, timestamps, etc.), and double checked with our own ability to identify the dogs. 

\begin{figure}
 \centering
 \begin{tabular}{cc}
  \includegraphics[width=0.4\columnwidth]{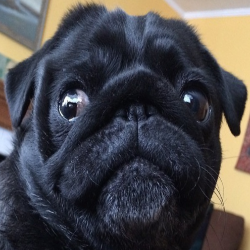}  & \includegraphics[width=0.4\columnwidth]{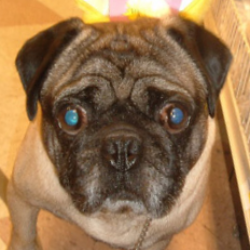} \\
  \includegraphics[width=0.4\columnwidth]{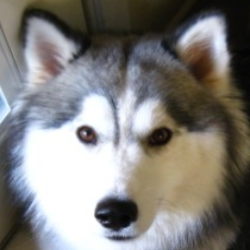} & \includegraphics[width=0.4\columnwidth]{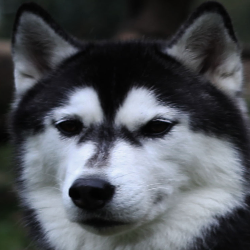} \\
  \end{tabular}
%
%
 \caption{A sample of Flickr-dog dataset. We acquired 374 photos licensed under Creative Commons from Flickr, representing 2 breeds (pugs and huskies), 21 individuals per breed, and at least 5 photos per individual. The choice of breeds intended to reflect a difficult case (pugs) and an easy one (huskies).}
 \label{fig:Flickr-dog}
\end{figure}

Altogether, the Flickr-dog dataset has 42 classes and 374 photos. Figure~\ref{fig:Flickr-dog} shows a typical sample.

The annotation is very laborious and is the limiting factor in expanding the dataset. Still, while keeping the data acquisition manageable, we strove to make the dataset as challenging as possible. By focusing on just pugs and huskies, we prevented the classifier from simply identifying the breeds. By cropping the faces, we reduced background information that could give unfair clues to the classifier. The choice of 21 individuals per breed was not accidental either: it makes random matches just unlikely enough for the usual significance of 95\%.

\subsubsection{Snoopybook}

\begin{figure}
 \centering
 \begin{tabular}{cc}
  \includegraphics[width=0.4\columnwidth]{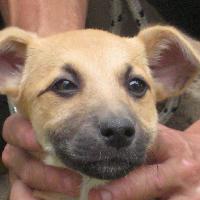} & \includegraphics[width=0.4\columnwidth]{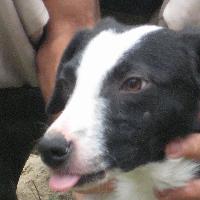} \\
 \end{tabular}

 \caption{A sample of Snoopybook dataset. This dataset has 18 mongrel dogs, with at least 5 photos per individual, for a total of 251 photos. With a less controlled array of individuals, Snoopybook is complementary to the well-controlled Flickr-dog dataset.}
 \label{fig:snoopybook}
\end{figure}

This dataset has 18 mongrel dogs --- mostly puppies --- with, again, at least 5 photos per individual, for a total of 251 photos. Each photo was registered to put the eyes and snout at the same position, and then resized to $200\times200$ pixels (Figure~\ref{fig:snoopybook}).

The Snoopybook dataset is complementary to Flickr-dog, as it offers a less controlled array of individuals. 

\subsection{Protocol}
\label{sec:protocol}

\begin{figure}[htb!]
 \centering
 \includegraphics[width=\textwidth]{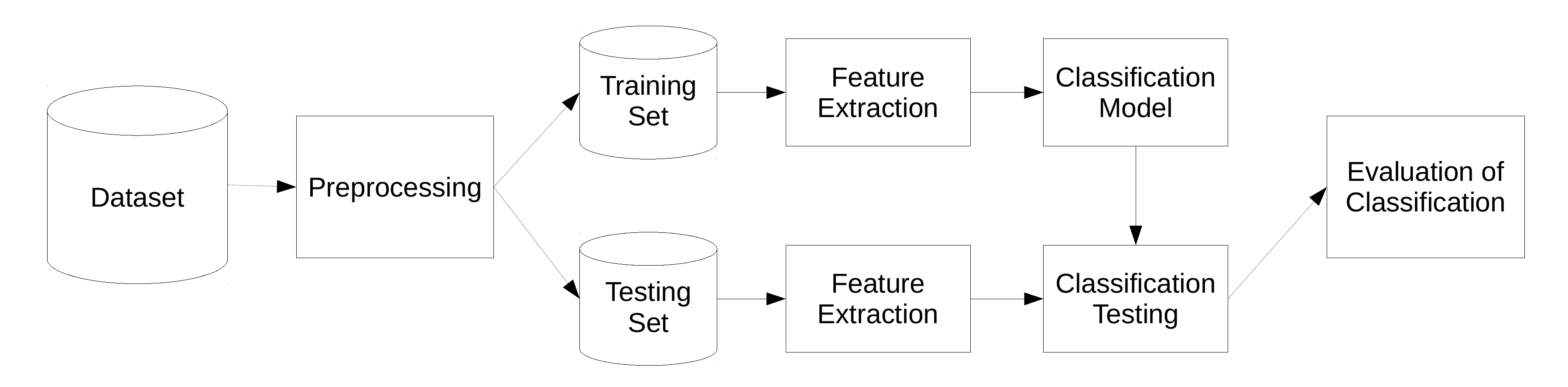}
 \caption{General pipeline of the experiments.}
 \label{fig:general_pipeline}
\end{figure}

Figure~\ref{fig:general_pipeline} shows the basic pipeline used in the experiments of this work.

The protocol was a stratified $k$-fold cross-validation, that splits the dataset into $k$ folds, preserving as much as possible the class proportions among the folds. We used 10 folds, with nine folds for training, and one for testing.

Our main metric is the balanced average accuracy, which is the arithmetic mean of the accuracy for each of the classes. We also employ confusion matrices for detailed analyses of the results.

For the retrieval experiment, we employ a top-$k$ recall, which ranks the classification scores for all classes, and counts the test as successful if the right class is among the highest $k$ scores.

\subsection{Techniques}

\subsubsection{Baselines}
\label{baselines}

As baselines, we employed the human face recognition algorithms readily available in OpenCV~\cite{opencv}: LBPH, EigenFaces and FisherFaces. OpenCV offers end-to-end algorithms for a facial recognition that start with the image and end with the class label; we chose to use those algorithms without modification.  EigenFaces' best results employed 80 components, while FisherFaces kept all components (corresponding, in our experiments, to the number of classes/individuals). LBPH most accurate and agile results used the standard eight points on a circle of radius one.

We also employed a sparse method for face recognition~\cite{xu2011}, This method is based in two phases, in which the first phase reconstruct a test image by the training subset of images, and the second phase also reconstructed the test image, but only using a fourth of the nearest faces, to define the contribution of class to the reconstruction.

\subsubsection{Shallow CNNs --- BARK}
\label{original1}


We propose Convolutional Neural Networks as the most promising solution for dog facial recognition. We follow Pinto et al.~\cite{Pinto2011} by optimizing the network architecture, and employing random weights. The weights themselves are not optimized. The resulting network extracts features that are fed to a linear SVM. 

The network structures explored during optimization follow a scheme defined in Cox et al.~\cite{Cox2011}. The network consists mainly of stacked layers, going from one up to three layers with the same hyperparameter space, which on the bottom of the architecture there is a pre-processing normalization of the input and at the top a linear SVM. Each of the intermediary layers have the following components, in this order: 

\begin{enumerate}
 \item \textbf{Convolution Linear Filters}: Bank of linear filters to be applied in the input signal with convolution operation.
 All kernels have random values respecting a uniform distribution;
 \item \textbf{Activation Function}: Function responsible for clipping the activation to a parametric range.
 \item \textbf{Pooling}: Neighbors filter activations will be pooled together, spatial down sampling even further the output.
 \item \textbf{Normalization} (optional): Normalization of the pooled output based on the magnitude of the response of neighbors within some region.
\end{enumerate}

A detailed description on the functions' mathematics of each component of the network can be found in Cox et al.~\cite{Cox2011}.

We implemented the solution with simple-hp\footnote{www.github.com/giovanichiachia/simple-hp}, a high-level wrapper for Bergstra's Hyperopt-convnet~\cite{Bergstra2013}. 
Simple-hp employs SVM from scikit-learn 0.14.1, with linear kernel and parameter $C = 1e5$. 
A total of 2000 architectures were evaluated using both the Random and the Tree of Parzen Estimators (TPE) optimizers~\cite{Bergstra2013}.
The hyperparameters space for optimizing the network architecture comprised: 
\begin{itemize}
 \item \textbf{Input size (in pixels)}: 64x64, 128x128, 256x256, original image size;
 \item \textbf{Number of layers in network}: 1, 2, 3;
 \item \textbf{Normalization after pooling}: Yes / No;
 \item \textbf{Filters in each layer (amount)}: 32, 64, 128, 256;
 \item \textbf{Size of the filters (pixels)}: 3, 5, 7, 9;
 \item \textbf{Exponent of pooling layers (scalar)}: 1, 2, 10;
 \item \textbf{Stride of pooling layers (pixels)}: 1, 2, 4, 8.
\end{itemize}



\begin{figure}[htb!]
 \centering
 \includegraphics[width=\textwidth]{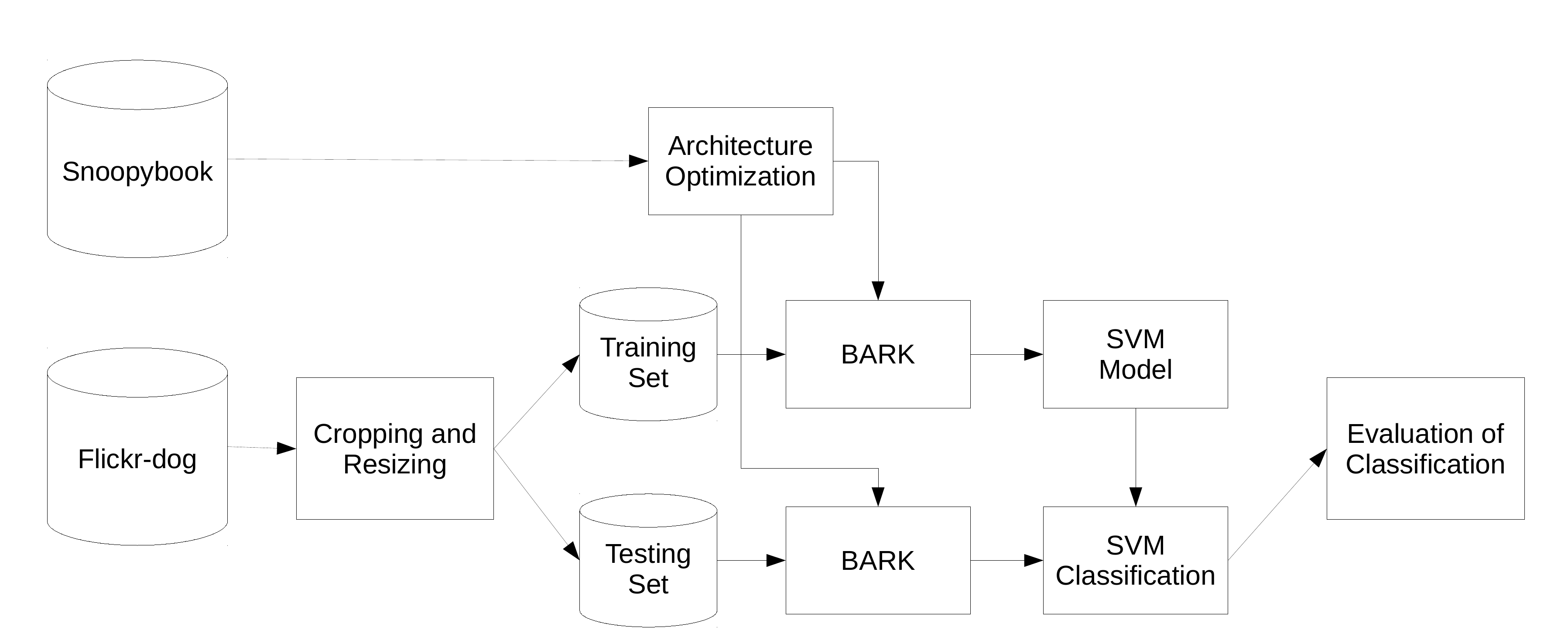}
 \caption{An example for the BARK pipeline.}
 \label{fig:bark_pipeline}
\end{figure}

In summary, this first proposal employs rather shallow convolutional networks whose architecture/hyperparameters are trained, but whose weights are kept random. We called it BARK --- Best Architecture to Retrieve K9. Figure~\ref{fig:bark_pipeline} shows an example of pipeline for the BARK proposal.

\subsubsection{Deep CNNs --- WOOF}
\label{original2}

%


The second original proposal employs off-the-shelf OverFeat~\cite{Sermanet2013} as Convolutional Neural Network. 
Sermanet et al.~\cite{Sermanet2013}, proposed the OverFeat model similar to Krizhevsky et al.~\cite{Krizhevsky2012}, improving the network design and inference step. 
The Krizhevsky et al.~\cite{Krizhevsky2012} network's architecture comprises eight layers, divided in five convolution-based and three fully-connected layers. The authors emphasize the following as most important aspects of this model, presented in order of importance: Rectified Linear Units (ReLUs) neurons, which allow much faster learning than \textit{tanh} units; Local Response Normalization, for aiding in generalization; Overlapping Pooling, that slightly lowers error rates by hindering overfit. Also, for reducing overfitting, data augmentation techniques and dropout~\cite{Hinton2012} were used.

Sermanet et al.~\cite{Sermanet2013} network won the localization task of ImageNet 2013 challenge, and achieved 14.20\% top-5 error rate on the classification task, against Krizhevsky et al.~15.30\%.
This error rate was obtained with an 8-layer architecture -- five convolutional and three fully-connected -- with some modifications. ReLUs non-linearities are still present in the convolutional layers, along with max pooling. However, there are three different aspects: Local Response Normalization is not used; there is no overlapping in pooling regions; and it was set a smaller value for stride on the 1st and 2nd layer, resulting in larger feature maps.
A more accurate version of the network was also designed, improving their results in ImageNet about two percentage points. In this model, there are six convolutional and three fully-connected layers. Besides the extra convolutional stage, there are other variations, mainly on the strides and sizes of the first layer and feature maps from all convolutional stages. With these variations, the number of connections almost doubled.


As before, the network is applied extracting features that are fed to a linear SVM. We resize all input images to $221\times221$ pixels, as required by OverFeat. The OverFeat package offers two models: \emph{accurate} or \emph{fast}. We pick the former, in virtue of its better performance. We used the libSVM 3.17 with linear kernel and default parameters ($C = 1.0$).

\begin{figure}[htb!]
 \centering
 \includegraphics[width=\textwidth]{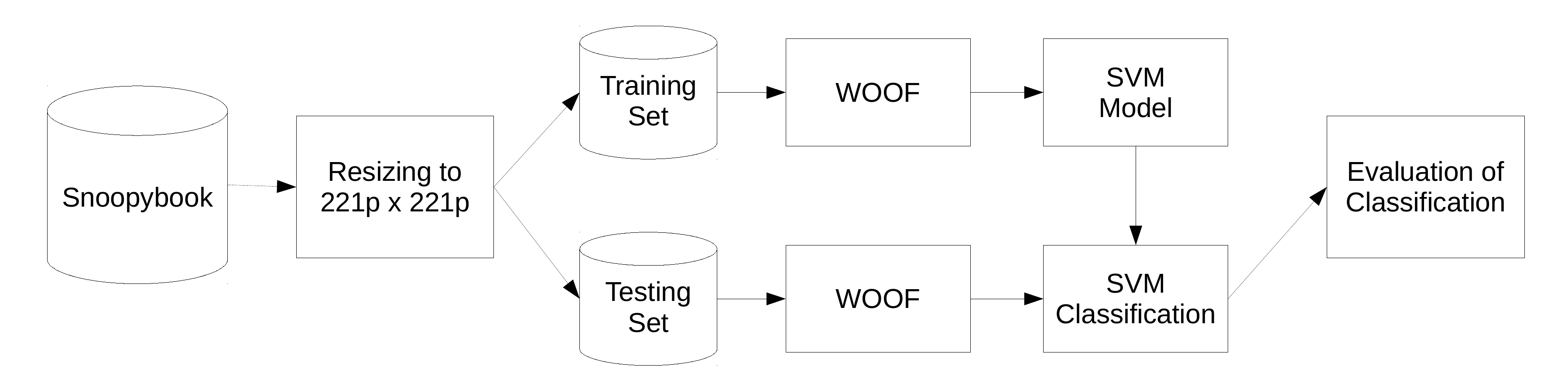}
 \caption{Pipeline example of the WOOF proposal applied to the Snoopybook dataset.}
 \label{fig:woof_pipeline}
\end{figure}

In summary, this second proposal employs very deep, but pre-trained, convolutional networks. 
No training or fine tuning was performed with Deep CNNs because it would be infeasible given the handicap in the number of samples from datasets utilized.
Only the SVM is trained. We called this second proposal WOOF --- Wields Off-the-shelf OverFeat Features. Figure~\ref{fig:woof_pipeline} shows an example of the WOOF proposal applied to the Snoopybook dataset.

\section{Experiments}
\label{sec:experimental}


We understand that, in real scenarios, the dog search range may be narrowed by information readily known to the dog's owner and shelter keepers, but our experiments concern the descriptor effectiveness in general conditions.

We proposed three questions to guide our work: \textit{Do human face recognizers generalize to canine faces?}, \textit{Are general-purpose convolutional networks competitive with special-purpose facial recognizers?}, and \textit{How accuracy behaves as the number of retrieved individuals vary?}.

\textit{Do human face recognizers generalize to canine faces?} To answer that question, we evaluate our baselines, ready-to-use human facial recognizers available in OpenCV and the sparse method in both our datasets. 

Table~\ref{tab:results} shows the balanced average accuracy obtained with four human face recognizers: EigenFaces~\cite{Sirovich1987}, FisherFaces~\cite{Belhumeur1997}, LBPH~\cite{Ahonen2006}, and Sparse~\cite{xu2011}. However, all four off-the-shelf human facial recognizers have rather poor results for dogs, while neural networks performed significantly better. The best hand crafted result, in Flickr-dog dataset, was LBPH's 43.2\%, and the general purpose network yielded 66.9\%. For Snoopybook, the sparse representation had the best result of 60.5\% and the network, 89.4\%.
That answers our first question: dog facial recognition is not a trivial extension of human facial recognition.

\begin{table}[h!]
 \centering
 \caption{Balanced accuracies (in \%) for all evaluated methods in both datasets. Off-the-shelf human facial recognizers (first four lines) tend to work poorly. WOOF got the best overall results, but BARK with the architecture trained in Flick-dog was a good second best. As we expected, Flickr-dog was more challenging than Snoopybook, due to presence of just two breeds.}
 \large
 \begin{tabular}{l|c|c}
  \multicolumn{1}{c|}{\multirow{2}{*}{ Methods} } & \multicolumn{2}{c}{Dataset} \\
  \cline{2-3}
  & Flickr-dog & Snoopybook \\
  \hline
  EigenFaces~\cite{Sirovich1987} & 33.9 & 41.6 \\
  FisherFaces~\cite{Belhumeur1997} & 22.7 & 55.4 \\
  LBPH~\cite{Ahonen2006} & 43.2 & 56.1 \\
  Sparse~\cite{xu2011} & 39.9 & 60.5 \\
  \hline
  $\textrm{BARK}_{\textrm{flickr}}$ & \textbf{67.6} & 81.1 \\
  $\textrm{BARK}_{\textrm{snoopy}}$ & 49.1 & 64.4 \\
  WOOF & \textbf{66.9} & \textbf{89.4} \\
  \hline
 \end{tabular}
 \label{tab:results}
\end{table}

\textit{Are general-purpose convolutional networks competitive with special-purpose facial recognizers?} To answer that, we evaluate two original solutions based upon existing Convolutional Networks: BARK and WOOF. BARK has two training steps: one for the network architecture, and one for the SVM. The dataset used to train the architecture appears in the subscript $\textrm{BARK}_{dataset}$ while the dataset used for training the SVM appears at the table header in Table~\ref{tab:results}. We ensure that the test set excludes all samples used in any  training.

During BARK architecture optimization, many different combination of the hyperparameters of the network were evaluated by extracting the features from the training data and feeding it to the SVM in a cross-validation manner, with different sub-splits from the training dataset in train and validation.
The mean accuracy of the SVM was used to determine the best architecture found, and then this architecture, and only this, was evaluated with the test data.

Table~\ref{tab:results} shows the results in the bottom three lines. As explained, 
in $\textrm{BARK}_{\textrm{flickr}}$, Flickr-dog is used for training the architecture, and in $\textrm{BARK}_{\textrm{snoopy}}$, Snoopybook is employed.


BARK and WOOF performed similarly on Flickr-dog, but WOOF performed better on Snoopybook. We believe that the difference in the latter case is due to the deeper, and pre-trained, OverFeat network employed in WOOF. OverFeat is a very deep network, pre-trained with millions of images, thousands of which contain dogs.

Curiously, $\textrm{BARK}_{\textrm{flickr}}$ performed better for both datasets. This might be due to Flickr-dog having more data and with individuals more similar to each other, demanding more discriminative features, thus leading to a better optimization of the architecture.

What lead us to believe that the greatest impact on the results did not come from the parameters, and hyperparameters space, employed here, which seems to be sufficient for a near optimal performance, considering this methodology.
Rather, the performance rely more on a appropriate training dataset, with challenging cases so the optimization can focus on the most discriminative features, and with distinct breeds, since different breeds may have different features that help discriminating between its individuals.

\textit{How accuracy behaves as the number of retrieved individuals vary?}. To answer the third question, we evaluate the best technique's --- WOOF's --- top-$k$ recall as we vary $k$.  That corresponds to a retrieval scenario reflecting the real-world application: there is a database of labeled dog pictures, the user comes with one photo of an unknown dog, and the system provides a few best matches for the user to consider.

\begin{figure*}[t!]
    \centering
    \begin{tabular}{c}
     \includegraphics[width=0.75\textwidth]{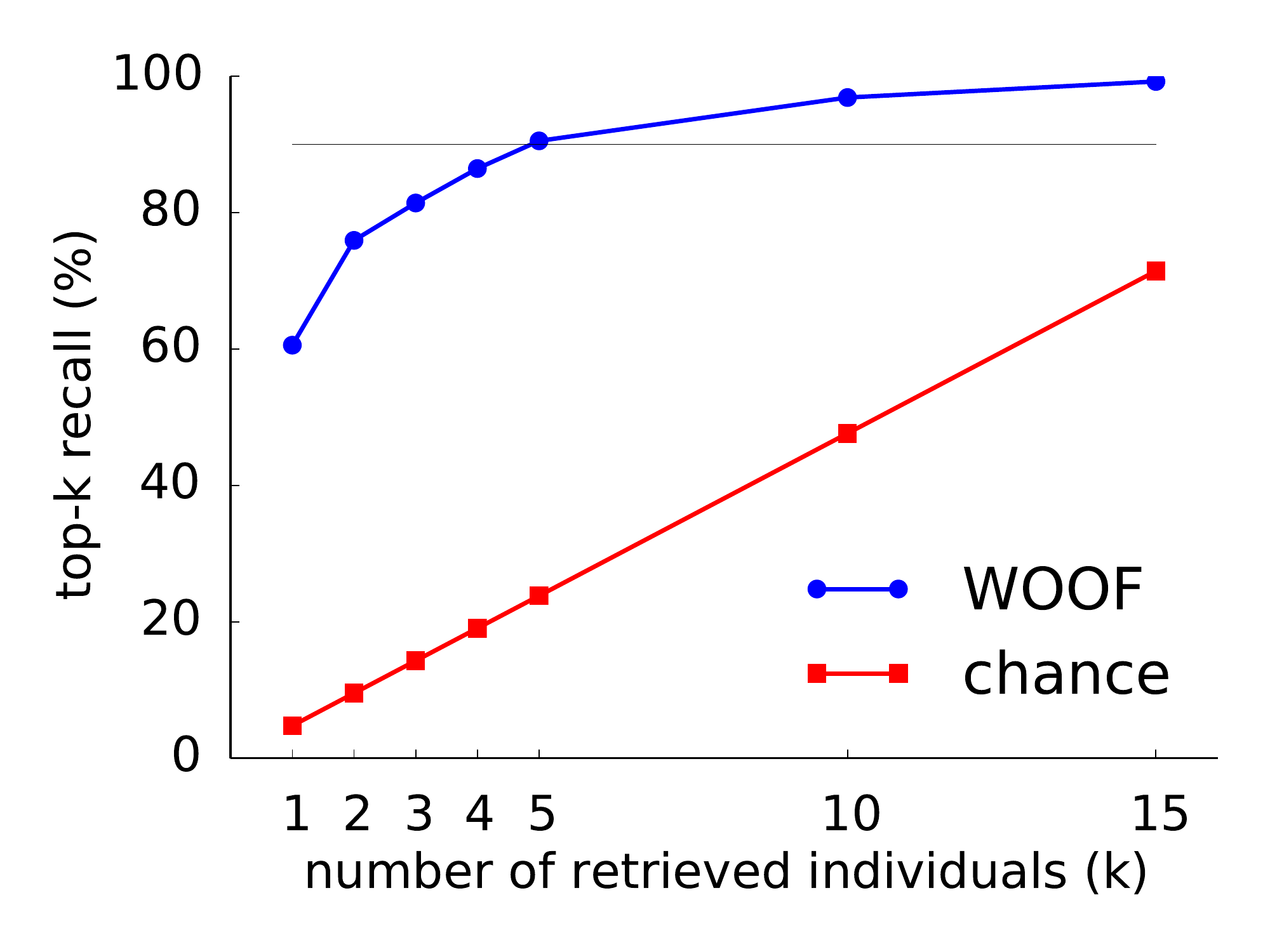} \\
     \includegraphics[width=0.75\textwidth]{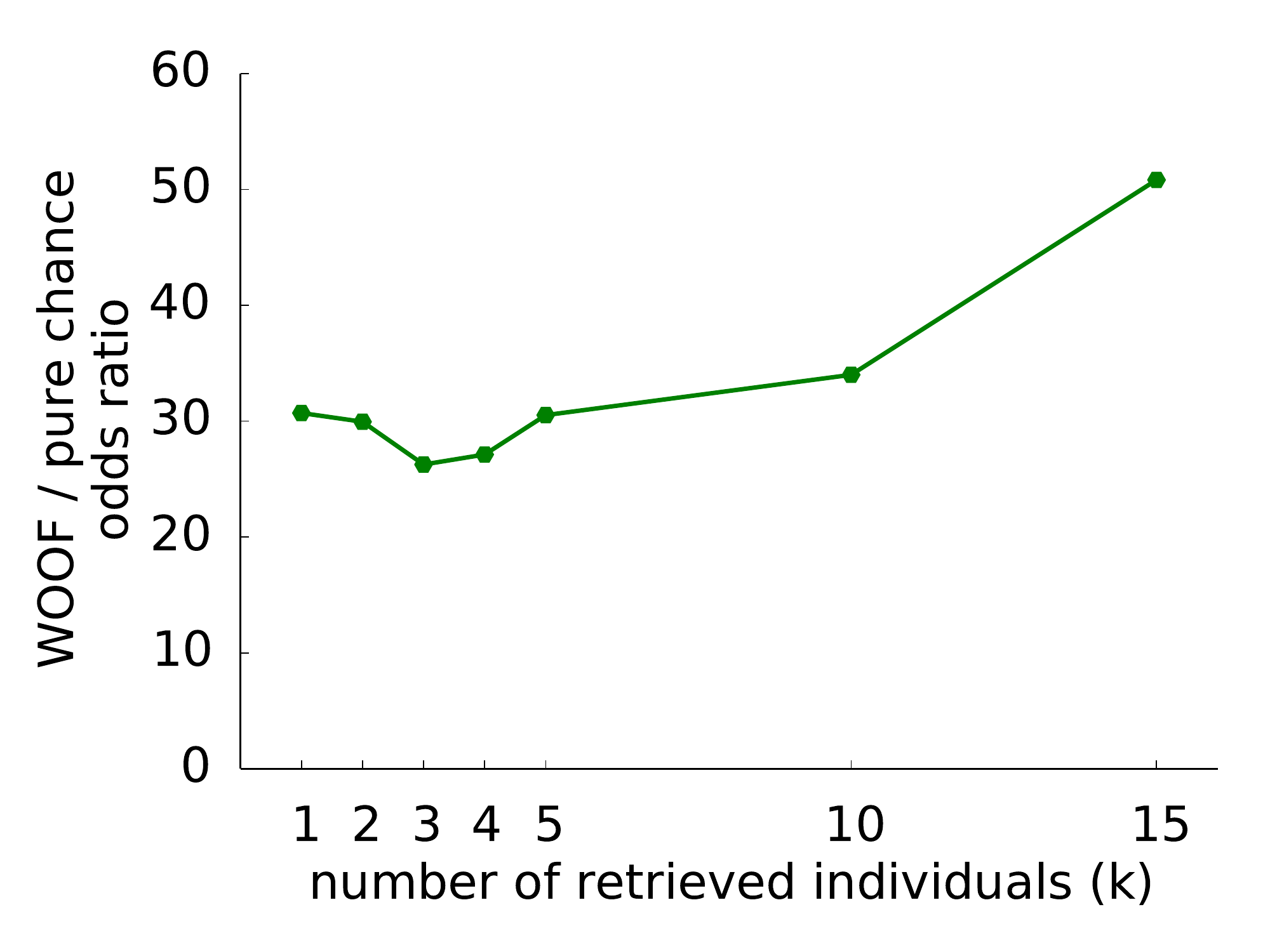} \\
    \end{tabular}

    \caption{Above: Top-$k$ recall (whether the correct class was among the $k$ most probable predicted classes) for WOOF in Flickr-dog. For $k = 5$ recall is already above $90\%$. The red line shows the expected (computed) recall for pure chance. Below: The recall for all $k$ stays well above random chance, with the odds-ratio actually improving for large $k$. }
    \label{fig:retrieval}
\end{figure*}

We present the top-$k$ recall (explained in Section~\ref{sec:protocol}) for $k = \{1, 2, 3, 4, 5, 10, 15\}$ in Figure~\ref{fig:retrieval}--Above. For $k = 5$, top-$k$ recall is already over $90\%$. In order to evaluate the robustness of WOOF with a limited-size dataset, we made a strict comparison with random chance. The odds ratio is a standard way to compare two probabilities, fractions or rates, dividing the odds of the first by the odds of the second. For a given probability, fraction or rate $p$, $\textrm{odds}(p)=p/(1-p)$. We used $k/21$ as random chance, more stringent than the $k/42$ that would allow for confusion between the breeds in Flickr-dog. Still, WOOF performs well above it for all tested values of $k$, as we show in Figure~\ref{fig:retrieval}--Below. 

\begin{figure}[h!]
 \centering
 \begin{tabular}{c}
  \includegraphics[width=\textwidth]{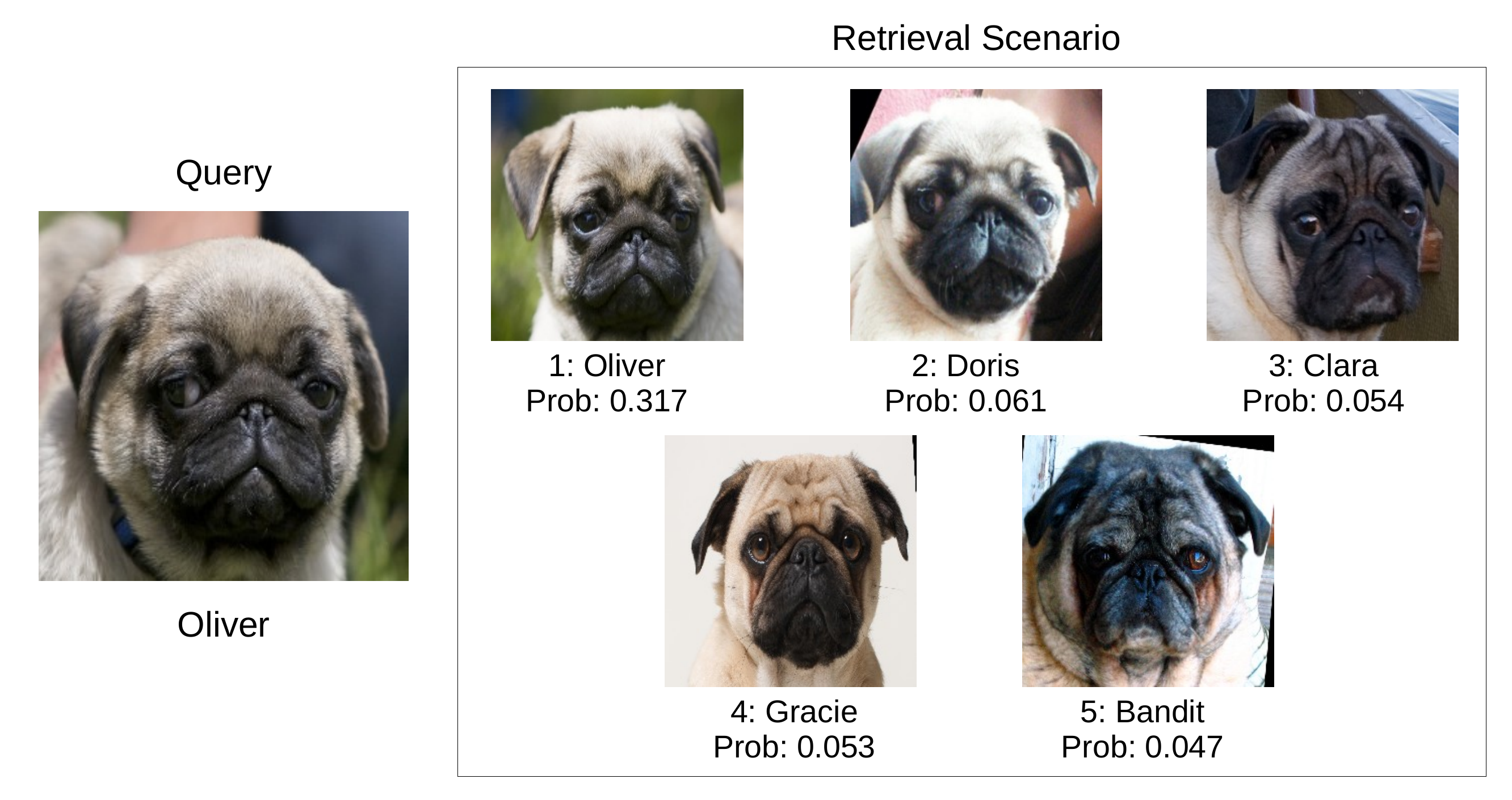} \\
  \includegraphics[width=\textwidth]{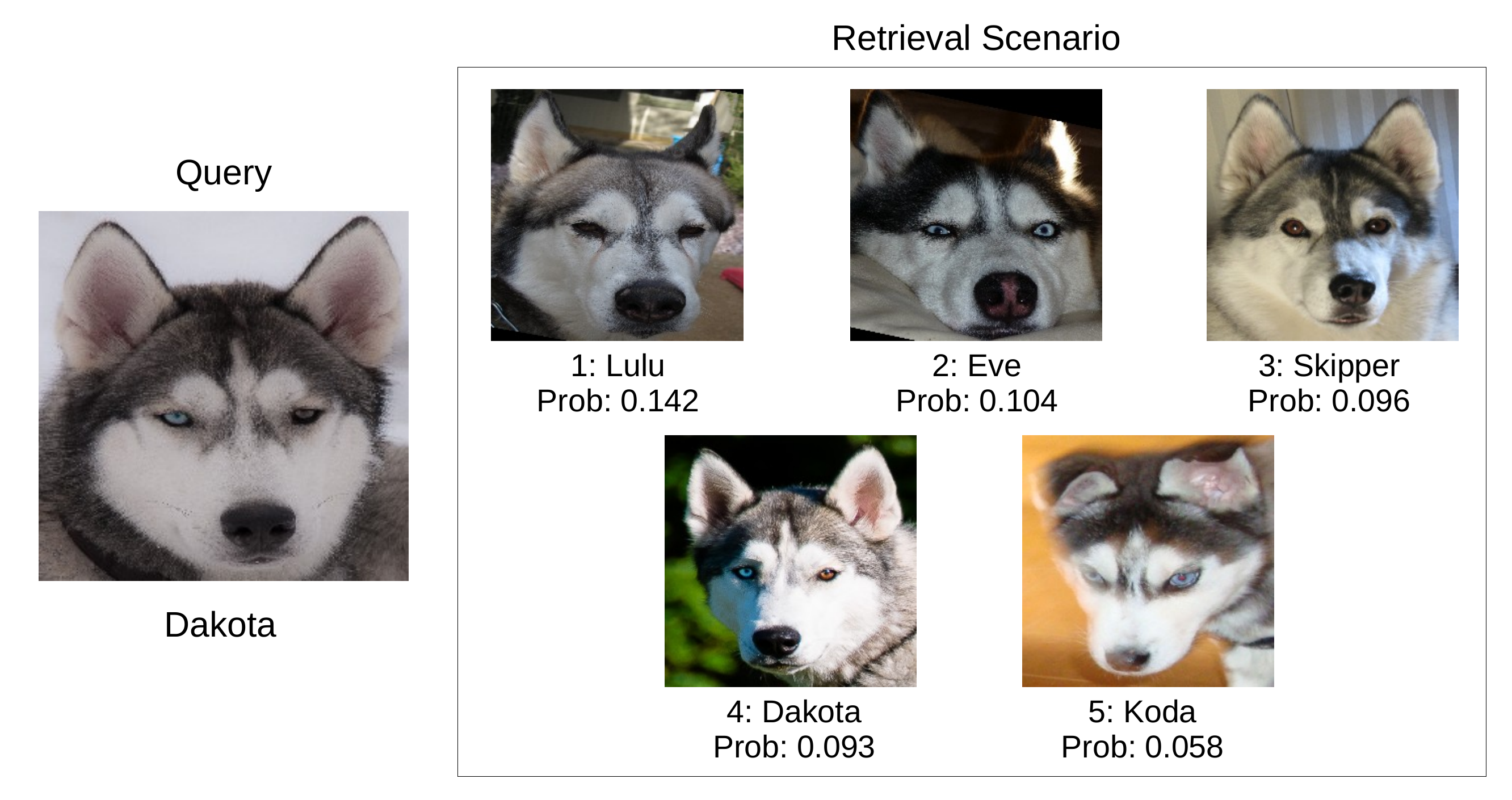}
 \end{tabular}

 \caption{Retrieval example for an example query from each one of the breeds. Above: Retrieval using a Pug dog as query, correct predicting its class. Below: Retrieval using a Husky dog, in which it predict correctly only in the fourth retrieved class.}
 \label{fig:retrieval_example}
\end{figure}

To complement the result above, we show two examples in a retrieval scenario in Figure~\ref{fig:retrieval_example}.

Figure~\ref{fig:retrieval_example}--Above presents an example of a correct predicting of the Oliver query in the first probable class, with a large margin for the second probable class. Figure~\ref{fig:retrieval_example}--Below does not predict the correct class for the Dakota query until the fourth probable class. However, the probabilities of the first class and the fourth class are too close, showing that our approach is really robust, as shown by Figure~\ref{fig:retrieval}--Above.

\begin{figure}[h!]
 \centering
 \includegraphics[width=\textwidth]{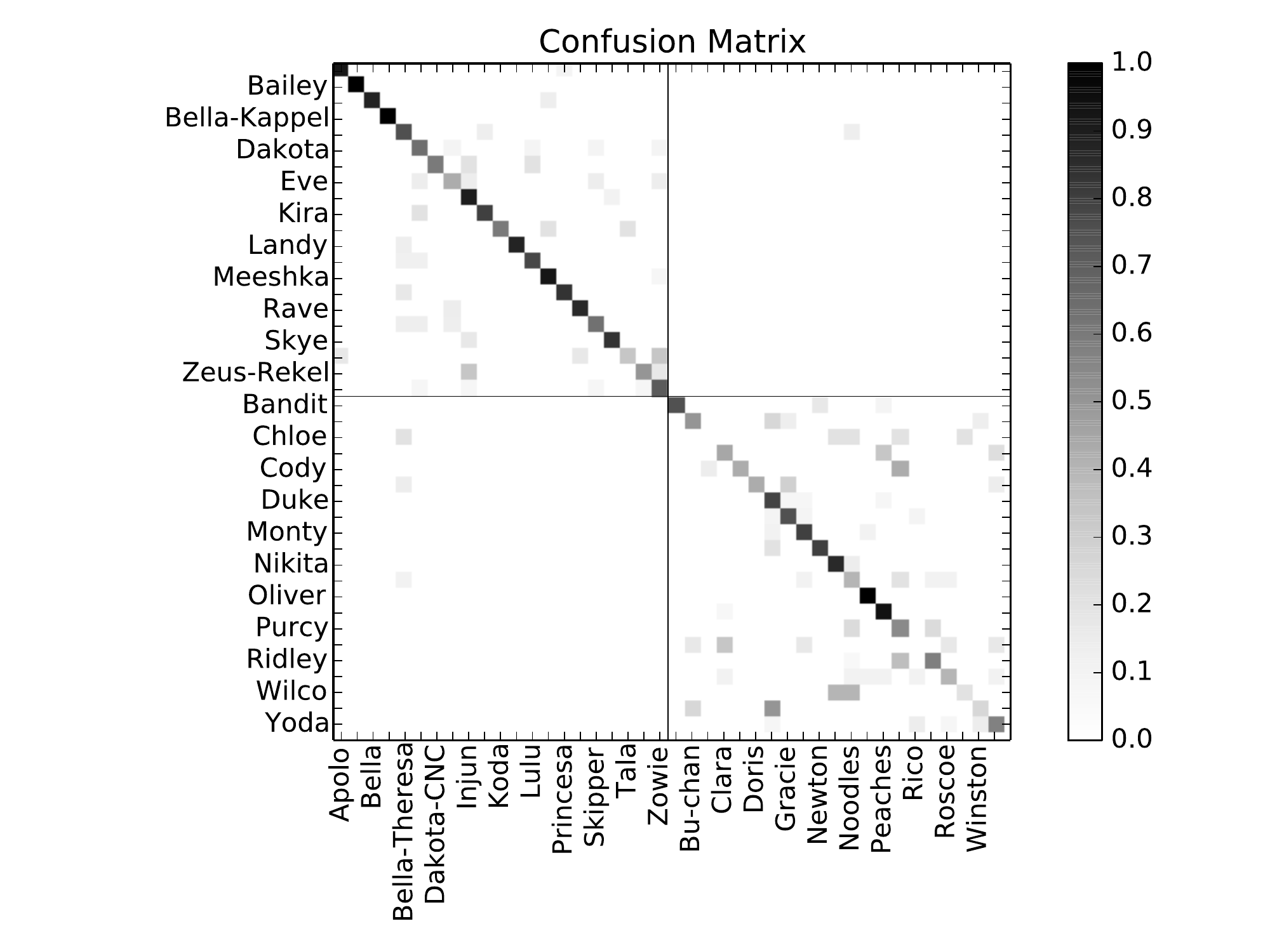}
 \caption{Confusion matrix for WOOF in Flickr-dog: huskies at top-left (Apolo to Zowie), pugs at bottom-right (Bandit to Yoda). 
 The two breeds appear clearly separated: almost all mistakes were between dogs of the same breeds. As expected huskies were easier to identify. Accuracy for huskies only was 75.14\%; for pugs, only 54.38\%.}
 \label{fig:confusionMatrix}
\end{figure}

We also provide a more in-depth profile of WOOF, with a confusion matrix (Figure~\ref{fig:confusionMatrix}). The matrix shows that very few mistakes mix-up dogs of different breeds. As expected, almost all mistakes are between dogs of the same breed.
Considering only huskies, accuracy was 75.14\%; while for pugs, accuracy was 54.38\%. That confirms our expectation that pugs would be harder to identify than huskies.


Neural networks are known to have issues with overfitting, mainly with small datasets, which is the case with our sets, specially Snoopybook. This does not apply to the WOOF method, since it is a descriptor based upon a general purpose network, pre-trained over a completely different set. As for BARK, we trained networks on both datasets, so when it is optimized for a set and tested on another, it is also not overfitted. In the last case, where network and SVM training is made in the same set, we avoid overfitting by isolating a subset for each phase.

Moreover, a form of assessing if a network is overfitted is applying it over different data. This is done for BARK, as shown in Table~\ref{tab:results}. The network optimized for Flickr-dog and tested on Snoopybook yielded good results -- worse than WOOF, but considerably better than handcrafted descriptors. Snoopybook, being a very small set, produced a weaker network. Nevertheless, its results on Flickr-dog are still better than LBPH.

\section{Conclusions}
\label{sec:conclusions}


We advanced dog identification from a new perspective, using facial features. To the best of our knowledge, our work is the first to address the problem, as existing art mostly sees animals as groups, not as individuals. Although ecologists have growing interest in identifying individual wild animals, pet identification is a new frontier for computer vision.

Our main motivation is to evaluate if pet biometry can be useful for retrieving lost animals, complementing existing solutions like dog tags, tattoos and micro-chips, that require special forethought, or may be lost.


We introduced two new datasets annotated with individual dog labels: Flickr-dog and Snoopybook. Data acquisition was the bottleneck for growing those datasets, as the manual annotation is very laborious. However, while keeping data acquisition manageable, we have shown that facial recognition in dogs is possible with accuracies much above pure chance. We hope that this work will spark the interest of other groups, and allow more aggressive, collective, efforts of data acquisition.




The performance of off-the-shelf classical Human Facial Recognizers was rather poor for dogs, showing that dog facial recognition is not a trivial extension of human facial recognition. Convolutional Networks, both shallow and deep, proved more successful. A very deep pre-trained OverFeat network --- WOOF --- showed the best results, with the shallow network with special architecture optimization --- BARK --- achieving a very decent second best. The performance was very promising, especially when contrasted to the literature that shows that dog recognition is hard even for human dog experts, with 81\% accuracy in optimal conditions~\cite{diamond1986faces}.

Our results also suggest margin for improvement with more training samples per individual. This suggests that for the optimal operation of a real-world system, owners should be stimulated to register as many pictures as possible.

We foresee applications beyond finding lost pets. Ecologists have growing interest in identifying individual wild animals, to detect migration patterns. There is also a forensic interest, in the case of stolen horses and cattle. We would like to address those exciting use cases, extracting features from the rest of the animal, not only the face. For certain animals, the coat can provide clues even more distinctive than the face, but because animals are deformable objects, getting invariant enough features is very challenging.





\section*{Acknowledgments}

Special thanks to the veterinary doctor Marjorie de Oliveira Franco,
along with the Zoonoses Control Center of S\~ao Jos\'e dos Campos - SP,
for producing the Snoopybook dataset, and providing it for research.
And thanks to Giovani Chiachia for its help on using simple-hp, tips on how to perform the experiments and helping organizing the dataset.


\bibliographystyle{spbasic}      
\bibliography{refs}   


\end{document}